\documentclass[11pt,a4paper]{article}
\usepackage[hyperref]{acl2020}
\usepackage{times}
\usepackage{latexsym}
\usepackage{graphicx}

\usepackage{microtype}

\aclfinalcopy % Uncomment this line for the final submission
\title{Instructions for ACL 2020 Proceedings}

\author{First Author \\
  Affiliation / Address line 1 \\
  Affiliation / Address line 2 \\
  Affiliation / Address line 3 \\
  \texttt{email@domain} \\\And
  Second Author \\
  Affiliation / Address line 1 \\
  Affiliation / Address line 2 \\
  Affiliation / Address line 3 \\
  \texttt{email@domain} \\}

\date{}

\begin{document}
\title{KddRES: A Multi-level Knowledge-driven Dialogue Dataset for Restaurant Towards Customized Dialogue System}

\author{
Hongru Wang$^{1}$, Min Li$^2$, Zimo Zhou$^3$, Gabriel Pui Cheong Fung$^1$, Kam-Fai Wong$^1$ \\
$^{1}$The Chinese University of Hong Kong\\
$^{2}$Communication University of China \\
$^{3}$Fudan University\\
hrwang@se.cuhk.edu.hk, kfwong@se.cuhk.edu.hk\\
}
\maketitle
\begin{abstract}

Compared with CrossWOZ (Chinese) and MultiWOZ (English) dataset which have coarse-grained 
information, there is no dataset which handle fine-grained and hierarchical level information 
properly. In this paper, we publish a first Cantonese knowledge-driven Dialogue Dataset for 
REStaurant (KddRES) in Hong Kong, which grounds the information in multi-turn conversations to one 
specific restaurant. Our corpus contains 0.8k conversations which derive from 10 restaurants with 
various styles in different regions. In addition to that, we designed fine-grained slots and intents 
to better capture semantic information. The benchmark experiments and data statistic analysis show 
the diversity and rich annotations of our dataset. We believe the publish of KddRES can be a 
necessary supplement of current dialogue datasets and more suitable and valuable for small and middle
enterprises (SMEs) of society, such as build a customized dialogue system for each restaurant. The 
corpus and benchmark models are publicly available.

\end{abstract}

\section{Introduction}
% task-oriented dialogue system
% Currently, only large enterprises can afford the very expensive chatbots. Since costs can be reduced significantly, SMEs can have their own chatbots, and they can re-allocate their existing resources to improve their service levels.
Task-oriented dialogue system is a system or chatbot which can help human to complete one predefined task(e.g. finding a restaurant or an attraction), which can be grouped into three classes, pipeline, end-to-end and between the above two types which some systems use joint models that combine some(but not all) of the four dialog components\cite{takanobu2020goaloriented}. Human-computer conversation becomes very important due to its promising alluring and potential commercial values\cite{Chen_2017}. However, there still is less application for small and middle enterprises (SMEs) than big companies because of the lack of large-scale high-quality dialogue data. Many corpus\cite{zhu2020crosswoz,budzianowski2018multiwoz,zhou2020kdconv,eric2019multiwoz,zang-etal-2020-multiwoz} before were collected to build a decathlon instead of specialist, which consist of multi-domain, cross-domain and try to handle various sub-tasks in one dialogue which is rare in real scenario, and some datasets\cite{wencond16,henderson-etal-2014-second} which focus on one domain ignores the fine-grained slots information and the complexity of real scenario.

% fine-grained data like price of meat and name of meat.
% constrains just one the user is assumed to want to know some information about this unique restaurant(i.e. name=Mcdonald's).
The first Chinese large-scale cross-domain task-oriented dataset (CrossWOZ), which contains 6k dialogue sessions and 102k utterances for 5 domains, including hotel, restaurant, attraction, metro and taxi\cite{zhu2020crosswoz}. This dataset focuses on hotel, restaurant and attraction, trying to handle cross-domain tasks like finding a hotel near-by the specific attraction and so on. Although it can provide invaluable suggestions and guidelines for tourists, it ignores fine-grained slots and values in a specific domain, which is a common problem for current corpus. For example, the restaurant domain in CrossWOZ ignores the price of dishes, location, operation time and so on. By contrast, \cite{budzianowski2018multiwoz} released a multi-domain English dialogue dataset spanning 7 distinct domains, it contains 10k dialogue sessions with annotation of system-side dialogue states and dialogue acts. But the follow-up work\cite{eric2019multiwoz,zang-etal-2020-multiwoz} indicates the presence of substantial noise in the dialogue state annotations and dialogue utterances and then fixed some annotation errors. All datasets put more attention on building a task-oriented dialogue system, which can handle composite tasks from different domains.

To handle the aforementioned problems, we propose the first Cantonese knowledge-driven dialogue dataset for restaurant (KddRES). The elaborate experiments and data analysis shows that our data has a more complicated data format and captures more information which more likely happened in a real scenario not only by the complexity of Cantonese but also the fine-grained slots information. Figure 1 shows a few turns of a dialogue in KddRES.
Our contributions can be grouped into three folds.  

\begin{itemize}
\item We published one knowledge-driven dialogue dataset for restaurant (KddRES), as far as we know, this is the first Cantonese dialogue dataset for task-oriented dialogue system.
\item We also provide natural language understanding baselines and the result shows the current pre-trained language model still can not get satisfactory performance in our dataset.
\item Exhaustive data analysis and experiments show our dataset is more diverse and 
complicated because of various and reasonable data formats existing in real scenarios, which 
indicate the potential capability to develop a customized dialogue system for a specific 
restaurant.
\end{itemize}

\begin{figure}
\centering
\includegraphics[scale=0.5]{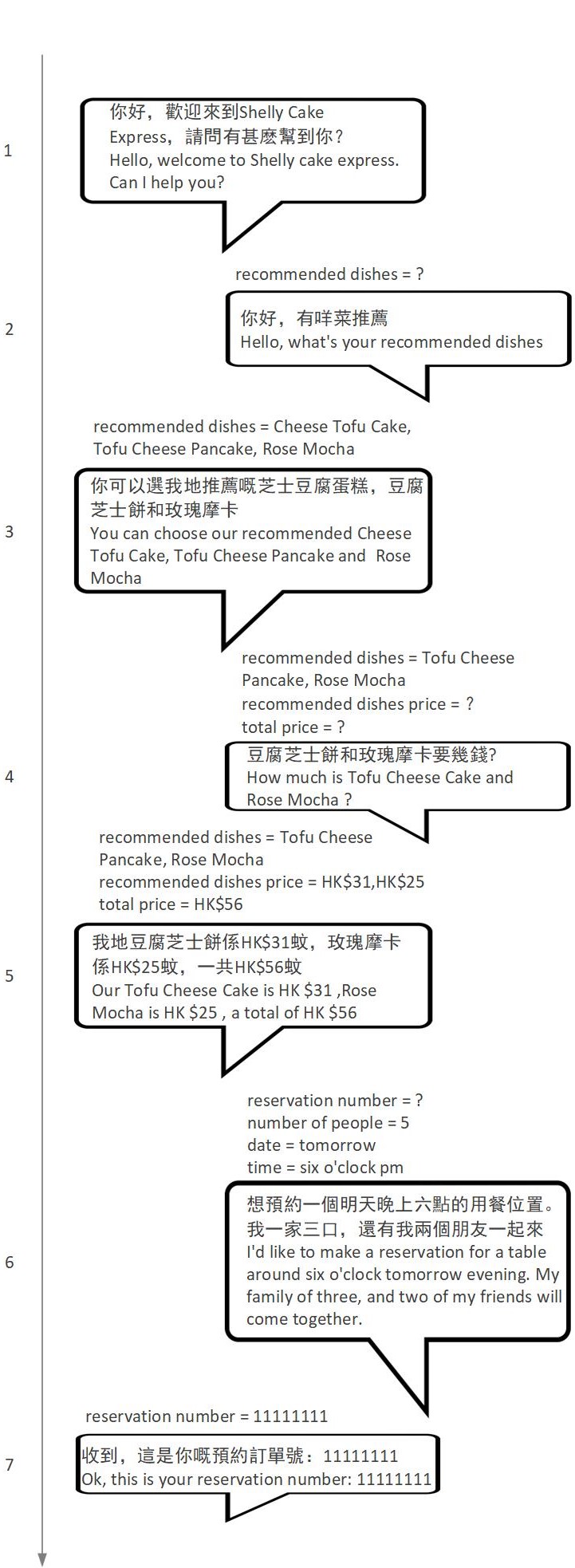}
\caption{A few turns of a dialogue in KddRES.}
\end{figure}

\section{Related Work}
% limited data and knowledge-driven
% human-to-human human-to-machine
Task-oriented dialogue system attracts more attention since the application of virtual assistant Apple’s Siri\footnote{https://www.apple.com/ios/siri/}, Microsoft’s Cortana\footnote{https://www.microsoft.com/en-us/cortana/} and XiaoIce\cite{zhou2018design}\footnote{https://www.msXiaoIce.com/}, Google Assistant\footnote{https://assistant.google.com/}, and Amazon’s Alexa\footnote{https://developer.amazon.com/alexa/}\cite{shum2018eliza}. All virtual assistants come from big companies but small and middle enterprises (SMEs) also want to build their own dialogue system to help the customers. Due to a lack of corresponding datasets and tremendous investment, task-oriented dialogue system still can not be a usual and normal marketing tool for SMEs.

% english: dialog bAbI tasks data, 
To facilitate the development of conversational models, most of the existing corpus focus on multi-domain or cross-domain, mainly focus on hotel, restaurant and so on. Stanford dialog dataset\cite{eric2017keyvalue} contains 3031 dialogues which span schedule, weather and navigation. Another well-known dataset is derived from a series of competitions on the task of Microsoft dialogue challenge. \cite{li2018microsoft} released human-annotated conversational data in three domains(movie-ticket booking, restaurant reservation, and taxi booking). Until the appearance of the largest multi-domains dataset\cite{budzianowski2018multiwoz}, the changes of dialogue goals are allowed.

Lots of researchers also public some Chinese dialogue datasets recently. JDDC Corpus\cite{chen2019jddc} is a large-scale real scenario Chinese E-commerce conversation corpus with more than 1 million multi-turn dialogues. CrossWOZ\cite{zhu2020crosswoz} is the first Chinese cross-domain Wizard-of-Oz task-oriented dataset which encourages natural transition across domains in context. All aforementioned datasets which focus on multi-domains try to build a decathlon dialogue system, but ignore more fine-grained slots information and user requirements in one domain. For example, people may want to inquiry about the price of a dish, the location of the restaurant and the operation time when they finally decide to book a table 
but most of the current datasets ignore the requirement.

% Booking 
As for specific restaurant domain, DSTC2\cite{henderson-etal-2014-second} is a human-to-machine restaurant booking dataset which used for dialogue states tracking, bAbI\cite{bordes2016learning} tasks data is more commonly used in end-to-end dialogue system because of lack of annotation information. \cite{wen2016networkbased} introduced CamRest676 which has a set of coarse dialogue acts for each user turn.

Compared with previous datasets in restaurant, our dataset has rich annotations and can be 
used for both task-oriented dialogue system and end-to-end system. To best of our knowledge, 
this is the first Cantonese knowledge-driven dialogue dataset for restaurant in Hong Kong, we
hope the publication of this dataset can be the catalyst of appearance of customized dialogue
system for SMEs.

\section{Data Collection}
The dialogue database simulates the real scenario where users in Hong Kong seek restaurant information and make a booking at one specific restaurant, which is the biggest difference of our corpus. User and System, two roles in a dialogue session, exchange values of different slots with each other, and make annotations of actions and states for each dialogue explicitly. The data collection process is outlined as follows:
\begin{itemize}
\item Database Construction: We first choose 10 representative restaurants with different styles in Hong Kong from Web to build the database. The corresponding slots are extracted from their information. And some new slots are added to better imitate the real scenario like waiting time and table size.

\item Goal Generation: The goal generator was designed based on the restaurant information database. A dialogue goal, usually as a user goal, contains two types of slots: informable slots and requestable slots. The former indicates the user's preference and information that user need to inform the system, but the later one requires the user to ask the system for the information. For example, an inform slot, such as time=``18:00'', means the user wants to have a dinner at 18:00 which he needs to inform the system, and a request slot, like price=``?'', request the user to ask this information. To make it easier for workers to understand the goal, we elaborated templates to form natural language task descriptions.

\item Dialogue Collection: A website was developed to collect dialogue data, which randomly assigns two roles to workers: user and system, to simulate real-life restaurant consultation dialogues. In the dialogue, users request information from systems or inform something to them according to the task description, and the system responds accordingly. 

\item Dialogue Annotation: Annotations are divided into dialogue actions and states. There are different types of dialogue actions: general, request, inform, info-confirm and so on. States are user states and system states, which record the change process of semantic forms on the user and system side. Workers are required to make annotations for their own messages before sending them. Finally, each dialogue contains a structured goal, task description and a message history (includes user state/system state, action, and message text). After that, four experts are hired to double-check the dialogue action, state and other elements to ensure the quality of the dialogue database.

\end{itemize}

\subsection{Database Construction}
% different level of data 
10 different styles Hong Kong restaurants are selected from Facebook\footnote{https://www.facebook.com/} and OpenRice\footnote{https://www.openrice.com/en/hongkong} to build the database. The corresponding slots are extracted from this information. However, this information can only cover a few scenarios in real life, in order to better imitate the real scenario, some new slots are added. For example, reservation number, current waiting number, current waiting time,  and table size are added as new slots to simulate the scene of the customer's reservation. When a user wants to make a reservation, he needs to provide the number of people, date and time of his meal plan to the chatbot, and get a reservation number from the chatbot.  Table 1 shows the comparison of slots between the restaurant realm of CrossWOZ\cite{zhu2020crosswoz} and our dataset KddRES. Compared with CrossWOZ\cite{zhu2020crosswoz}, KddRES has more slots and can cover more scenarios. Besides, CrossWOZ\cite{zhu2020crosswoz} consists of many domains and focuses more on  multi-domain or cross-domain which ignores  the  fine-grained  slots' information  in  one domain. KddRES improves this by introducing the secondary slots.

\begin{table}[ht]
    \centering
    \begin{tabular}{l} % {|l|c|r|} left居左显示，c居中显示，r居右显示
    \hline
    \textbf{Common}\\
    name, rating, cost, recommend dishes\\
    address, phone, \textbf{open time}\\
    \hline
    \textbf{Crosswoz(restaurant)}\\
    nearby attract, nearby rest, nearby hotels\\
    \hline
    \textbf{KddRES}\\
    characteristic, \textbf{dishes}, subway, bus\\
    \textbf{table size},number of comments\\
    \textbf{current waiting number}\\
    \textbf{current waiting time}, \textbf{take out}\\
    querying for number support\\
    reservation support, \textbf{parking}, discount\\
  	\hline
    \end{tabular}
    \caption{All slots in Crosswoz(restaurant) and KddRES (translated into English). Slots in bold means this slot of KddRES can be divided into secondary slots shown in Table 2}
    
\end{table}

A secondary slot refers to a slot that containing more detailed information. For example, when a user wants to know about a restaurant's dishes, task-oriented dialogue system can not only tell the name of the dishes, but also the price of the dishes. User can also specifically ask about the price of a certain dish. Dishes and Dishes-price are the secondary slots in this scenario. Table 2 shows the slots with this feature in KddRES.

\begin{table}[ht]
    \centering
    \scalebox{0.8}{
    \begin{tabular}{l|l} % {|l|c|r|} left居左显示，c居中显示，r居右显示
    \hline
    Primary slots& Secondary slots\\
    \hline
    dishes& dishes-name\\
    &dishes-price \\
    \hline
    open time&open time-date\\
    &open time-time\\
  	\hline
  	table size&table size-name\\
  	&table size-size\\
  	\hline
  	current waiting number&current waiting number-table name\\
  	&current waiting number-number\\
  	\hline
  	current waiting time&current waiting time-table name\\
  	&current waiting time-time\\
  	\hline
  	take out&take out-support\\
  	&take out-time\\
  	\hline
  	parking&parking-support\\
  	&parking-price\\
  	\hline  	
    \end{tabular}}
    \caption{All secondary slots and their primary slots in KddRES(translated into English)}

\end{table}

\subsection{Goal Generation}
To generate a reasonable dialogue goal, we firstly classify all slots into four groups 1) basic information 2) location(subway and bus) 3) dishes and discount 4) additional information. There are different types of slots in these groups. And then we will generate the sub-goal with different probability $P$ for each group like 0.6 for group 1 but 0.4 for group 4. Also, we require the goal to meet some constraints. For example, the dialogue goal must have one slot in each of the first three groups and one in the last group. We encourage the user to change the goal during the session if there is no-offer or user wants to know more details and the changed goal is also stored.

\subsection{Dialogue Collection}
A  website was developed specifically which allows two persons to mimic different roles for dialogue collection. It allows two workers to talk synchronously and exchange their own message. To ensure workers are trained well, operating instructions and video demonstrations are provided in the website. After workers fill on the necessary personal information, like their name, they can freely choose the role: user or system and start a conversation. 

\subsubsection{User Side}
There are three components on the user side: task description, slot-value box, and dialog box. The task description is the natural language description of the dialogue goal. It tells the user what information needs to request and gives some constraints, such as the date and time of booking a table. The slot-value box is a table that every row combines a checkbox, slot, and value. The user selects a checkbox to mark dialogue action and fills in the value to update the user state. The user synchronizes the dialogue with the system in the dialog box. In each conversation, the user needs to 1) write a message based on the system reply. 2) make annotations in the slot-value box before sending a message,  and submit them. 3) terminates the dialogue if the dialogue goal is completed.

\subsubsection{System Side}
Compared with the user side, the system side has no task description, and the slot-value box adds a column of restaurant information. In each round, the system needs to 1) make appropriate responses in natural language based on previous user conversations and restaurant information. 2) make annotations in the slot-value box before sending a message. If there is no offer to the user's inquiry, the system will try to make a recommendation.

\subsection{Dialogue Annotation}
The whole processing of dialogue annotation is divided into two stages. At the first stage, the two roles in a dialogue session already explicitly choose the slot and fill the value during dialogue collection. Secondly, we used some rules to fine-tune the result from the first stage. Compared with CrossWOZ\cite{zhu2020crosswoz}, we introduce one new intent names \textbf{info-confirm}, this intent is designed for some special scenario like  the user wants to know whether or not the restaurant has this dish.

\section{Data Statistics}
A dialogue is considered an incomplete dialogue when the goals in a dialogue are not all achieved. After removing these incomplete dialogues, we collect 832 dialogues in total, the dataset is divided randomly into training set, validation set, and test set with a number of 600, 116 and 116. Table 3 shows the statistical indicators of the dataset. 
\begin{table}[h]
    \centering
    \scalebox{0.9}{
    \begin{tabular}{lccc} % {|l|c|r|} left居左显示，c居中显示，r居右显示
    \hline
    Statistical Indicators & Train & Valid & Test\\
    \hline
    Dialogues &600 &116 &116 \\
    Turns & 5795&1118&1109 \\
    Tokens  &97121&17758& 18232  \\
    Vocabs  &931&728& 705 \\
    Avg.goals per dialogue  &7.9 &7.8& 7.2  \\
    Avg.turns per dialogue  & 9.7&9.6& 9.6 \\
    Avg.tokens per turn  & 16.8 &15.9&  16.4\\
    Avg.dialogue acts per turn  &1.3 &1.2& 1.2 \\
    POSS  & 66.8\% & 66.7 \% & 74.7\% \\
  	\hline
    \end{tabular}}
    \caption{Data statistics, POSS is the proportion of dialogues that contain the secondary slots}
\end{table}

Dialogues and Turns refer to the total number of completed dialogues and utterances. Tokens refers to the number of all chars and vocabs refers to the number of vocabularies that have appeared in the dataset. The average goals and turns in a dialogue, tokens and dialogue actions in a turn are also calculated. We also calculate the proportion of dialogues that contain the secondary slots(POSS). POSS can reflect the ratio of detailed scenarios in the dataset. The POSS of KddRES in the training set, test set and validation set all exceed 60\%, which indicates that KddRES has a more detailed simulation of the real scenarios. The results indicate the diversity, delicacy and the ability of simulating real scenarios of our data.

Table 4 compares KddRES with other influential restaurant datasets. Because CrossWOZ\cite{zhu2020crosswoz} and MultiWOZ(restaurant)\cite{zang-etal-2020-multiwoz} pay more attention on the dialogues in multi-domain goal, the size of them is very large. Even so, the Avg.turns of Single-domain type dialogues in Crosswoz is only 6.8, less than kddRES. Meanwhile, the information in one specific domain is not detailed enough which limits the application. The average turns of KddRES is higher than CamRest676\cite{wencond16} and DSTC2\cite{henderson-etal-2014-second}. The average turns reflects the depth and breadth of a dataset. On this index, KddRES has reached a good level, indicating that KddRES has depth and breadth. In addition, Slots can reflect the richness of the scenario and context in a dataset. The number of slots in KddRES is much higher than the others, which indicates that KddRES contains rich contexts. The result of Table 4 indicates that our data reaches a good level both in quantity and quality. Also, these rich fine-grained slots bring many new challenges to NLU research in the following aspects: 
\begin{table*}[t]
    \centering
    \begin{tabular}{l|cc|ccc} % {|l|c|r|} left居左显示，c居中显示，r居右显示
    \hline
    Type & \multicolumn{2}{c|}{Multi-domain}& \multicolumn{3}{c}{Single-domain}\\
    \hline
    Dataset &CrossWOZ & MultiWOZ & DSTC2 & CamRest676  &KddRES\\
    \hline
    Language &CN & EN& EN&EN&CN(Cantonese) \\
    Dialogues &6012 & 10438&3235 &676 & \textbf{832}\\
    Turns &101626&132610&25488&1500 & \textbf{8022}\\
    Avg.turns &16.9 &13.7 &7.9& 2.2&\textbf{9.6} \\
    Slots & 72&25&8&3& \textbf{20}\\
  	\hline
    \end{tabular}
    \caption{Comparison of KddRES and other Influential restaurant datasets, the Avg.turns are for each dialogue}
\end{table*}

\begin{itemize}
\item Relation extraction: The introduction of fine-grained  slots make some slots separate into multi levels. For example, "dishes" is divided into two secondary slots: dish-name and dish-price. Figure2 shows the hierarchy of these slots. The values in these secondary slots are one-to-one corresponding. During the building of a dialogue system, especially the natural language understanding part. whether this kind of relation can be correctly extracted is important. For example, in the fifth turn of Figure 1, HK\$31 corresponds to the price of Tofu Cheese Cake, and HK\$25 corresponds to the price of rose mocha. It is not difficult for a human to understand this from the reply of system, but it's still difficult to make the system have the ability to extract this relation too. Besides, evaluating the ability of a dialogue system and building RL-based policy require establishing a non-human user simulator. In order to make the user simulator stable and reliable, how to slove this issue is also very meaningful.
\begin{figure}[h]
\centering
\includegraphics[scale=0.25]{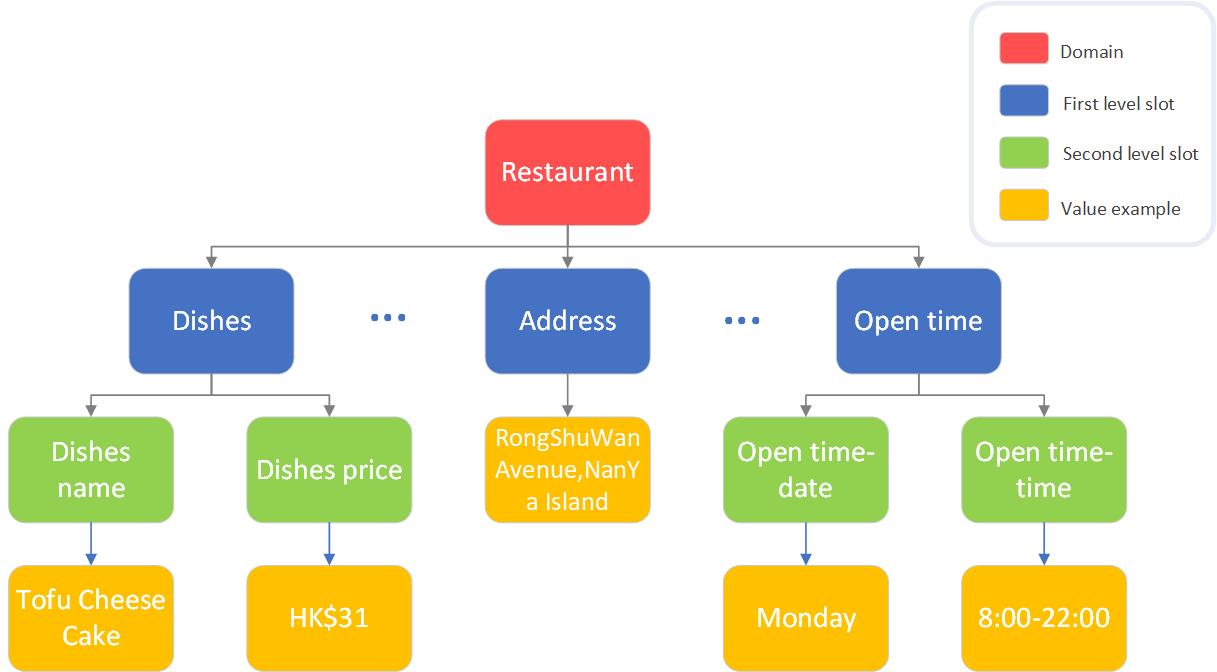}
\caption{The hierarchy of the slots(translated into English).}
\end{figure} 

\item Common sense knowledge: kddres provides many dialogues about asking for the total price of several dishes or recommended dishes and making an appointment. Like the fourth turn and sixth turn of Figure 1. In these scenarios, the system not only needs to accurately identify the name of the dishes (recommended dishes) inquired by the user, but also successfully infers that the total number of diners is five rather than three or two. For the first issue, NLU needs to ensure that the identified result is common sense. For example, in the fourth turn of Figure 1, how to effectively ensure that the identified dish name is "Tofu Cheese Cake" and "Rose Mocha" instead of "Tofu Cheese", "Rose" and "Mocha"; For the second issue, the chatbot can usually solve it based on some rules, for example, when NLU recognizes more than one number, the system will automatically do some simple calculation and give the total number. Apparently, the rule-based solution is too limited and inaccurate. An ideal dialogue system should respond in an empathic way rather than based on rules. It should infer the user's psychological state according to the events experienced by the user, without requiring the user to clearly explain their feelings. 
\end{itemize}

Exhaustive data statistics analysis shows our dataset have a more complicated data format and more suitable for customized dialogue system, which requires the model to capable to reason and retrieve correct answers. For example, most people will ask how long they need to wait to have a table. The model needs to find a suitable table size according to the number of persons to eat, and then find the waiting time.

\section{Experiment}
% baselines
Traditional task-oriented dialogue system is assembled from four components(Natural Language Understanding, Dialogue State Tracking, Dialog Policy and Natural Language Generation), but can not assure the best performance when assembling four best model in corresponding sub-task\cite{takanobu2020goaloriented},  Natural Language Understanding is the first part of a pipeline dialogue system, which takes an utterance as input and output its corresponding dialogue act. \cite{takanobu2020goaloriented} shows that a combination of model-based NLU and other rule-based components can get the best performance for task-oriented dialogue system, which indicates the importance of this task. NLU task can be usually divided into two sub-tasks, one is the intent classification, the other one is the slot filling. Intent classification is usually regarded as a classification problem to detect the user's intention. In contrast, Slot filling is usually regarded as a sequential labeling problem, the words in an utterance are assigned with semantic labels.

\subsection{Models}

\textbf{BiLSTM}
LSTM\cite{graves2013speech} model can better capture the long-distance dependence because LSTM can learn what to remember and what to forget through the training process. However, there is a problem when using LSTM to encode sentences: the information from back to front cannot be encoded. BiLSTM\cite{graves2013speech} is composed of forward LSTM and backward LSTM, which can better capture the bidirectional semantic dependency.

\textbf{BiLSTM-CRF}\cite{huang2015bidirectional}
can efficiently use past input features via a BiLSTM\cite{graves2013speech} layer and sentence level tag information via a CRF\cite{10.5555/645530.655813} layer. CRF\cite{10.5555/645530.655813} layer can add some constraints to the final predicted tags to ensure that the predicted tags are legal. With these constraints, the probability of illegal sequences in tag sequence prediction will be greatly reduced.

\textbf{BERTNLU}\cite{chen2019bert}
Pre-trained models like BERT\cite{devlin2018bert} have been proved to reach the state-of-art performance on lots of downstream tasks including natural language understanding\cite{zhu2020crosswoz}. BERTNLU uses BERT\cite{devlin2018bert} to joint training and complete the intent classification task and slot filling task.

\textbf{BERT-CRF}
Almost the same as BERTNLU\cite{chen2019bert} except an extra CRF layer to help improve the performance of the slot filling task.

\textbf{ERNIE}\cite{2019arXiv190409223S}
 is a pre-trained model like BERT\cite{devlin2018bert} but it also adds word segmentation information. Compared with BERT\cite{devlin2018bert}, ERNIE\cite{2019arXiv190409223S} learns the semantic relationships in the real world by modeling prior semantic knowledge such as entity concepts in massive data directly, which enhanced the semantic representation ability of the model. To use ERNIE\cite{2019arXiv190409223S} in NLU, we use the concat of token embedding, segment embedding and position embedding as the input of ERNIE\cite{2019arXiv190409223S} and use the output to joint train the intent classification task and slot filling task.

\subsection{Implement Details}
 BertAdam optimizer is used with the learning rate of $3*10^{-5}$ and epsilon of $10^{-8}$, The batch size is chosen as 20, and the hidden size is 768. ReLU\cite{agarap2018deep} is selected as the activation function. Before the output of pre-trained language model go to the next hidden layer, it will go through a dropout layer with a rate of $10^{-1}$ firstly. The model is trained for $4*10^{4}$ steps. 

\subsection{Result}
The number of results of intention classification and slot filling for one utterance is usually much less than all possible results. In this task We pay more attention to whether the identified intent and slot (positive examples) are accurate. meanwhile, Accuracy can not reflect the performance of a classifier when the distribution of positive and negative samples is unbalanced. Therefore, we choose F1, the harmonic average of precision and recall as the evaluation index of these models. Table 5 shows the result of these models on F1.

% our results
% our results
\begin{table}[ht]
    \centering
    \begin{tabular}{l|c|c|c} % {|l|c|r|} left居左显示，c居中显示，r居右显示
    \hline
    Model & In(F1) & Sl(F1)& Overall(F1) \\
    \hline
    BiLSTM & 90.69  & 91.95 & 91.37 \\
    BiLSTM+CRF & 91.89 & 91.65 & 91.76\\
    BERT  & 92.46 & 94.02  & 93.39 \\
    BERT+CRF & 92.03 & 94.34 & 93.42 \\
  	\hline
    \end{tabular}
    \caption{Experiment result}
\end{table}

Just like BERT's\cite{devlin2018bert} good capability in other NLP tasks, BERT\cite{devlin2018bert} also has a good result here. The F1 scores of Intent, Slot and Overall are all better than BiLSTM\cite{graves2013speech} and BiLSTM+CRF\cite{huang2015bidirectional}. However,we notice that ERNIE\cite{2019arXiv190409223S} has not achieved significant results. We think
the main reason is that the pre-trained corpus of ERNIE\cite{2019arXiv190409223S} only includes Simplified Chinese text, does not include Cantonese(Traditional Chinese) text. At the same time, these indicators show that KddRES has diversity and delicacy, which provides more challenges for the future research of task-oriented dialogue system.

\section{Conclusion and Future Works}
In this paper, we publicly release one multi-Level Knowledge-driven Dialogue Dataset for 
restaurant in real scenario with detailed data analysis and rerun competitive natural 
language understanding baselines on this dataset. All corpus and code will be available on 
the KddRES github. We hope the publication of this dataset can facilitate 
the development of customized dialogue system for SMEs. We left other components of 
task-oriented dialogue system like dialogue state tracking, policy learning to future work, 
and also we will expend our data to more domains.

\section{Acknowledgement}
The research described in this paper is partially
supported by HK GRF \#14204118 and HK RSFS
\#3133237

\bibliographystyle{acl_natbib}
\bibliography{acl2020}

\end{document}